\documentclass[letterpaper]{article} 
\usepackage{aaai23}  
\usepackage{times}  
\usepackage{helvet}  
\usepackage{courier}  
\usepackage[hyphens]{url}  
\usepackage{graphicx} 
\usepackage{natbib}  
\usepackage{caption} 
\usepackage{algorithm}
\usepackage{algorithmic}

\usepackage{amsmath}
\usepackage{amssymb}

\usepackage{multirow}
\usepackage{booktabs}
\usepackage{newfloat}
\usepackage{listings}
\DeclareCaptionStyle{ruled}{labelfont=normalfont,labelsep=colon,strut=off} 
\floatstyle{ruled}
\newfloat{listing}{tb}{lst}{}
\floatname{listing}{Listing}
\title{Rethinking Data Augmentation for Single-source Domain Generalization\\in Medical Image Segmentation}

\author{
    Zixian Su,\equalcontrib \textsuperscript{\rm 1, \rm 2}
    Kai Yao,\equalcontrib \textsuperscript{\rm 1, \rm 2}
        Xi Yang,\textsuperscript{\rm 2}\thanks{Corresponding author.}
    Qiufeng Wang,\textsuperscript{\rm 2}
    Jie Sun,\textsuperscript{\rm 2}
    Kaizhu Huang\textsuperscript{\rm 3}$^\dag$
}
\affiliations{
    \textsuperscript{\rm 1}University of Liverpool, Liverpool, the United Kingdom\\
    \textsuperscript{\rm 2}School of Advanced Technology, Xi'an Jiaotong-Liverpool University (XJTLU), Suzhou, China\\
    \textsuperscript{\rm 3}Data Science Research Center, Duke Kunshan University, Kunshan, China\\
    kaizhu.huang@dukekunshan.edu.cn, 
    xi.yang01@xjtlu.edu.cn
}
\usepackage{bibentry}

\begin{document}

\maketitle
\begin{abstract}
Single-source domain generalization (SDG) in medical image segmentation is a challenging yet essential task as domain shifts are quite common among clinical image datasets. Previous attempts most conduct global-only/random augmentation.  Their augmented samples are usually insufficient in diversity and informativeness, thus failing to cover the possible target domain distribution. In this paper, we rethink the data augmentation strategy for SDG in medical image segmentation. Motivated by the class-level representation invariance and style mutability of medical images, we hypothesize that  unseen target data can be sampled from a linear combination of $C$ (the class number) random variables, where each variable follows a location-scale distribution at the class level. Accordingly, data augmented can be readily made by sampling the random variables through a general form. On the empirical front, we implement such strategy with constrained B$\acute{\rm e}$zier transformation on both  global and  local (i.e. class-level) regions, which can largely increase the augmentation diversity.  A Saliency-balancing Fusion mechanism is further proposed to enrich the informativeness by engaging the gradient information, guiding augmentation with proper orientation and magnitude. As an important contribution, we prove theoretically that our proposed augmentation can lead to an upper bound of the generalization risk on the unseen  target domain, thus confirming our hypothesis. Combining the two strategies, our Saliency-balancing Location-scale Augmentation (SLAug) exceeds the state-of-the-art works by a large margin in two challenging SDG tasks. Code is available at \url{https://github.com/Kaiseem/SLAug}.

\end{abstract}

\section{Introduction}
Despite the success of deep learning applications in medical image analysis~\citep{med1,yao2022novel,dice}, many approaches are known to be data-driven and non-robust. 
However, it is often the case that the testing data exists distribution shift with the training data in terms of many factors, such as imaging protocol, device vendors and patient populations.

Many efforts have been made to alleviate the above-mentioned distribution shift problem. Generally, these methods either use extra data (e.g. Unsupervised Domain Adaptation~\citep{SIFA}) or require the diversity of the training data (e.g. \textit{Multi-source Domain Generalization}~\citep{wang2020dofe}). 
In this paper, we consider a more challenging but practical setting: single-source domain generalization (SDG)  in medical images. 
This task aims to learn a generalizable and robust  predictive model on the related but different  target domain with only single source domain data. 
The key to this problem lies in increasing the \textit{diversity} and \textit{informativeness} of the training data, i.e., data augmentation.

\begin{figure}[t]
\centering
\includegraphics[width=1\columnwidth]{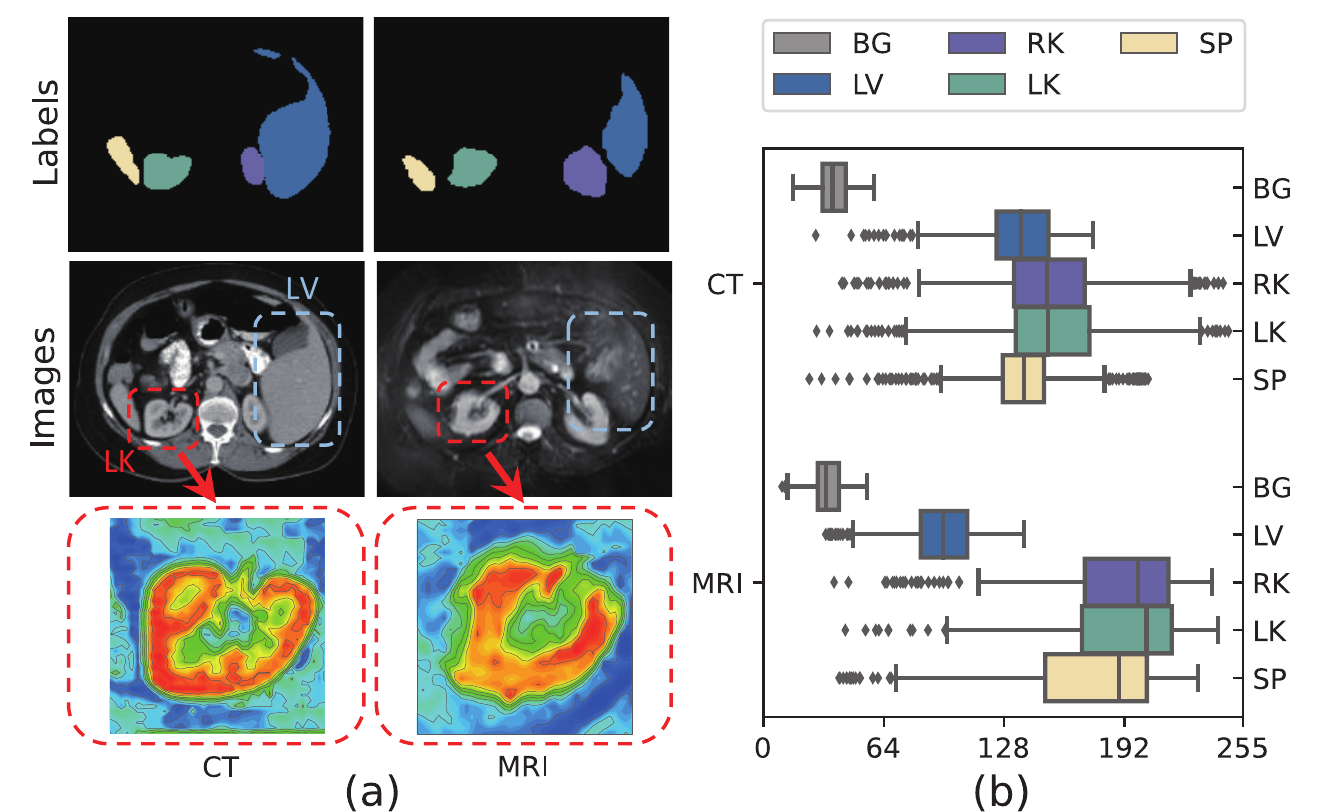}
\caption{Conceptual overview of our motivation. (a) Different domain images and partial enlarged views (colormap surface). (b) Box-plot of class-specific intensity value distribution across the overall domains of CT\&MRI. }
\label{fig:intro}
\end{figure}

Although previous augmentation methods have led to impressive results in generalization tasks, they  suffer from the following limitations in medical image segmentation: 1) Global-only augmentation performing transformation on the whole image limits the diversity of augmented images.
2) Random augmentation considers no constraint on the distribution of the augmented samples, which may lead to over-generalization and cause performance degradation.
3) General-task-specific augmentation specially designed for natural image is restricted from medical image tasks, such as weather  augmentation~\citep{volk2019towards} for automatic driving.
These limitations naturally motivate us to think:  \textit{How to improve augmentation in medical image segmentation?}
Meanwhile, by looking into medical image segmentation, there are three findings as follows:
1) {Class-level representation invariance.} Strong similarity exists in the pixel-level representation of the same class between different domains in medical images, as compared with natural images. For an illustrative example in Fig.~\ref{fig:intro}(a), the left kidney (LK) (red dashed square) shows the similar contrast tendency in both CT and MRI domains, i.e. they are relatively bright in the margins and dark in the center. 
2) Style mutability. Variance between domains may be reflected in the class-level overall contrast, texture, quality, and clarity. For example, if we compare the liver (LV) (blue dashed square) in Fig.~\ref{fig:intro}(a), it is obvious that this substructure is brighter compared with its surroundings in CT but blends into the background in MRI. This can also be observed through the blue box (LV) 
in Fig.~\ref{fig:intro}(b), whose mean value is close to other classes in CT but exists a gap in MRI. 
3) Segmentation task enjoys inherent advantages owing to pixel-level labels, meaning that the ground truth mask could be used to crop the class-level regions as a unit to perform augmentation. The combination of the class-level augmented samples are much more diverse.

Motivated from these findings, we argue that, in medical scenarios, it is reasonable to hypothesize that 
\textit{the unseen target data can be represented by a linear combination of $C$ (the class number) random variables. Each class-level random variable $x_U^c$ is sampled from a location-scale distribution. {That is to say, $x_U = \sum_{c=1}^{C} x_U^c $, where $x_U^c =  \alpha^c x_S^c + \beta^c$}. } It is noted that a location–scale distribution  belongs to a family of probability distributions  (e.g.  Normal distribution) parameterized by a location parameter and a non-negative scale parameter.

Based on this assumption, we propose a novel augmentation strategy, location-scale augmentation. The augmented data is randomly sampled from {the general form $x_S^{aug} =  \sum_{c=1}^{C}\alpha^c x_S^c + \beta^c$}, where $\alpha^c$ is the scale factor and $\beta^c$ is the location factor. 
With two individual modules: global location-scale augmentation (GLA) and local location-scale augmentation (LLA),
{this doubling augmentation strategy promotes the} diversity of the augmented samples exponentially  compared to the previous global-only augmentation.
On the empirical front,
our method performs constrained B$\acute{\rm e}$zier transformation on both global and local (i.e. class-level) regions.
GLA intends to immunize the model against  global intensity shift. LLA leverages the mask information to crop the class-level region and perform class-specific transformations, which aims to simulate the class-level distribution shift.  
Importantly, we  make a theoretical analysis in the setting of SDG and prove that the generalization risk can be indeed upper bounded by introducing the location-scale augmentation strategy, thus confirming our hypothesis.

Besides, {to address the over-generalization of random augmentation, }
we further devise a saliency-balancing fusion (SBF) strategy that engages the gradient distribution as an indicator for augmentation magnitude, making our augmentation process informative and oriented.
Specifically, we calculate the saliency map and 
optimize it from noisy to smooth to show the large-gradient area.  
Such regions in the GLA sample would be preserved, and then fused with LLA sample. {As such}, the augmented image would be neither too close to the original source distribution nor too far out-of-distribution. 
Experimental results validate that SBF could achieve a substantially lower source risk compared with the existing random/no fusion mechanism, leading to a tighter  generalization bound in our setting.

In summary, we make the following contributions:

1) To the best of our knowledge, we make a first attempt to investigate both global and local augmentation in medical image tasks. We propose a novel augmentation strategy, location-scale augmentation, that engages the inherent class-level information and enjoys a general form of augmentation.
Theoretically, we prove that the generalization risk of our augmentation strategy can be upper bounded.

2) We propose a saliency-balancing fusion strategy, providing augmentation with gradient information for proper orientation and intensity. Empirical investigations show  saliency-balancing fusion achieves lower source risk,  leading to a tighter bound {in overall generalization risk}.

3) Combining the two strategies, our Saliency-balancing Location-scale Augmentation (SLAug) for SDG  achieves superior performance in two segmentation tasks, demonstrating its outstanding  generalization capability.

\begin{figure*}[t]
\centering
\includegraphics[width=1.8\columnwidth]{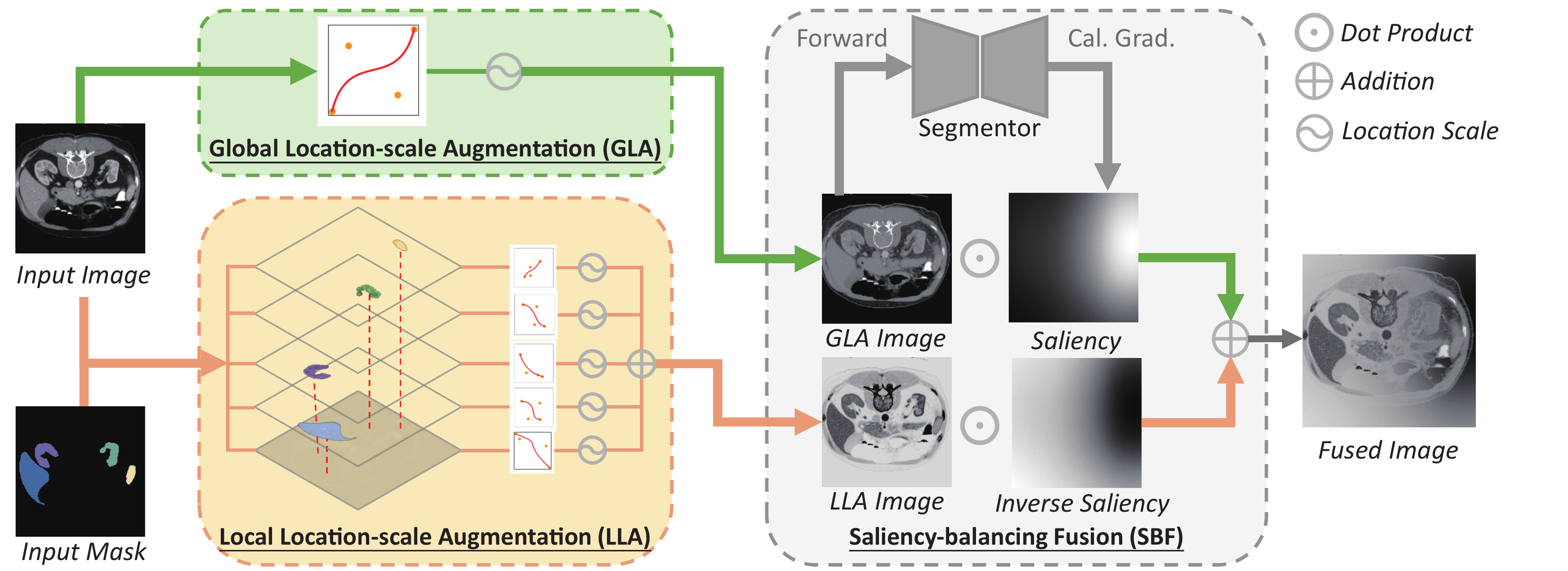}
\caption{{Workflow of the proposed Saliency-balancing Location-scale Augmentation (SLAug).} }
\label{fig:overall}
\end{figure*}

\section{Related Work}

\paragraph{Single-source Domain Generalization} Single-source Domain Generalization (SDG) aims to train a robust and generalizable predictive model against unseen domain shifts with only single source data, which is extremely challenging due to the insufficient diversity in training domain. Several methods~\citep{mixstyle,randconv,RSC} designed different generation strategies for augmentation in domain generalization tasks. Recently, in medical image fields, \citet{CSDG} used data augmentation to remove the spurious correlations via causal intervention, and \citet{dualnormal} proposed a dual-normalization based model to tackle the generalization task under cross-modality setting. Different from the previous works, we introduce additive class-level information into augmentation process, significantly boosting the generalization performance.

\paragraph{Mask-based Augmentation} Compared with traditional augmentation (random crop, flipping, etc.), DNN-orientated augmentation is designed from the learning characteristic view of DNNs. Recent efforts~\citep{classmix,objectaug,chen2021mask} have taken advantages of semantic mask to generate training samples. \citet{classmix} synthesized new samples by cutting half of the predicted classes from one image and then pasting them onto another image. \citet{objectaug} decoupled the image into the background and objects with the mask, and replaced the objects on the inpainted background image. However, the above methods are neither medical-orientated  nor designed for  domain generalization, making them not suitable in this medical segmentation  SDG scenario.

\paragraph{Saliency} Saliency information has been utilized in various fields of machine learning~\citep{wei2017object,kim2020puzzle}, which aims to figure out the region of interests in the images. Traditional methods~\citep{wang2014fast,hou2007saliency}  focused on exploring low-level vision features from images themselves. More recently, deep-learning based methods~\citep{zhao2015saliency,simonyan2013deep}  have been proposed for gradient-based saliency detection, which measures the impact of the features of each pixel point in the image on the classification result.  In this work, we follow  \cite{simonyan2013deep} to detect the saliency by computing gradients of a pre-trained model, which does not require modification on the network or additional annotation.

\section{Main Methodology}
This section details our proposed Saliency-balancing Location-scale Augmentation (SLAug). 
As illustrated in Fig.~\ref{fig:overall}, our framework consists of two parts: Location-scale Augmentation and Saliency-balancing Fusion.

\subsection{Preliminary}
In domain generalization tasks, we denote the training samples from the source distribution as $\mathcal{X}_S=\{(x_{i,S},m_{i,S})\}_{i=1}^{n_S}$ with total classes $C$, where $x_{i,S}$ denotes the $i^{th}$ training sample, $n_S$ is the number of training samples, $m_{i,S}$ denotes the corresponding groundtruth mask.  
Given an unseen
distribution $\mathcal{X}_U=\{(x_{i,U},m_{i,U})\}_{i=1}^{n_U}$, we aim to minimize
the error between prediction $\hat{m}_{i,U}$ and groundtruth mask
$m_{i,U}$  through a trained segmentor.

Given an image $x$ and its mask $m\in \{1,2,...,C\}^{W\times H}$, $m^c$ is defined as the binary mask of class c. $m^c_{(w,h)}=1$ means pixel $x_{(w,h)}$ falls in the region of class $c$.
The decouple of an image by the semantic class is represented as follows:
\begin{equation}
     x^c = m^c \odot x, \textstyle\sum_{c=1}^{C}x^c = x,
\end{equation}
where $x^c$ are the image regions belong to class $c$,   $\odot$ is dot product.  Note that since one pixel can correspond to only one class, the mask of one image is constrained as $\textstyle\sum_{c=1}^{C} m^c = \textbf{1}$, where \textbf{1} presents the all-ones matrix.

\subsection{Location-scale Augmentation}

We consider two types of augmentation,  Global Location-scale Augmentation (GLA), which increases the source-like images through global distribution shifting, and Local Location-scale Augmentation (LLA), which conducts class-specific augmentation to explore sufficiently diverse or even extreme appearance of unseen domain.

Different from previous works~\citep{randconv,CSDG}, which are based on generic assumptions and utilized random intensity/texture transformations, we leverage the monotonic non-linear transformation for medical images, as the absolute intensity values convey information of the underlying substructures \citep{buzug2011computed,forbes2012human}. Inspired by Model Genesis~\citep{zhou2019models}, we choose  Cubic B$\acute{\rm e}$zier Curve~\citep{mortenson1999mathematics}, a smooth and monotonic function, as our mapping function. {Specifically, we constrain our function by the the lowest and highest value of the non-blank image region. That is to say, given an image region with the lowest value $v_{\rm low}$ and highest value $v_{\rm high}$, the constrained B$\acute{\textit{e}}$zier function is given by: }
\begin{equation}
B\acute{\textit{e}}zier(t)=\textstyle\sum_{k=0}^3 (1-t)^{3-k}t^kP_k, t\in [v_{\rm low},v_{\rm high}],
\end{equation}
where $t$ is a fractional value along the length of the line. $P_0=(v_{\rm low},v_{\rm low})$ and $P_3=(v_{\rm high},v_{\rm high})$ are the start and end points to limit the range of value, and $P_1,P_2$ are the control points whose values are randomly generated from [$v_{\rm low}$,$v_{\rm high}$]. 
The inverse mapping function can be generated by swapping the start point and end point to $P_0=(v_{\rm low},v_{\rm high})$ and $P_3=(v_{\rm high},v_{\rm low})$. This non-linear transformation function is tagged as $\mathcal{F}_{\rm p}(\cdot)$, where ${\rm p}$ is the probability to perform the inverse mapping. 

\begin{figure}[t]
\centering
\includegraphics[width=0.95\columnwidth]{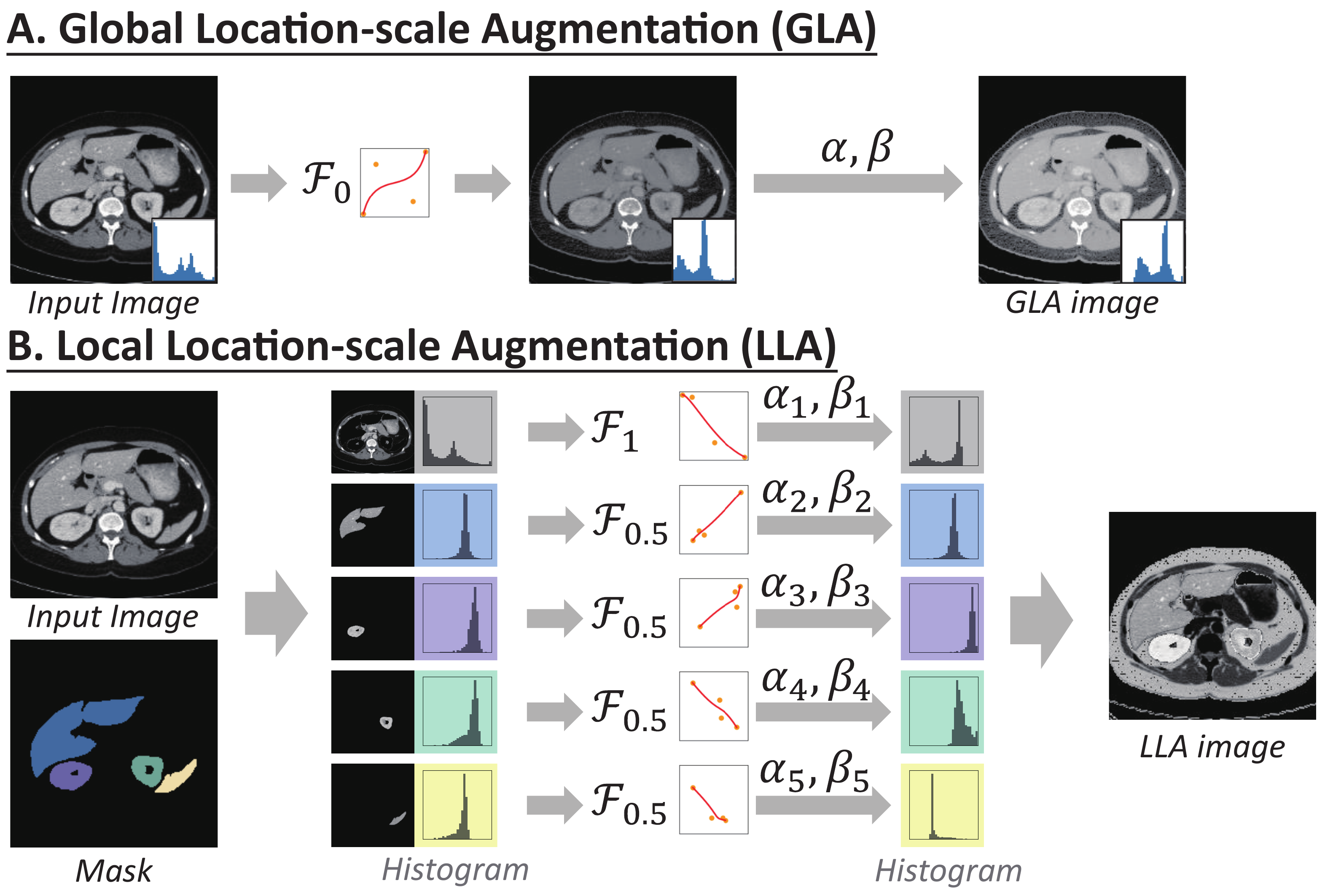}
\caption{Illustration of proposed global and local location-scale augmentation. $\mathcal{F}_{\rm p}$ denotes the non-linear transformation, while{ $\alpha$ and $\beta$ are location-scale factors}.}
\label{fig:bezier}
\end{figure}

Before augmentation, $x$ is min-max normalized in the range of [0, 1]. Then, GLA  performing global distribution shifting can be described as:
\begin{equation}
{\rm GLA}(x) = \alpha\mathcal{F}_0(x) + \beta,
\label{eq:gla}
\end{equation}
where $\alpha\sim\mathcal{TN}(1,\sigma_1)$,
$\beta\sim\mathcal{TN}(0,\sigma_2)$ are {location-scale factors}, and $\sigma_1$ and $\sigma_2$  are standard deviations of the two truncated Gaussian distributions respectively. By using $\mathcal{F}_0$, the probability of inverse is zero to ensure the GLA samples have  similar appearance with original images. 
In contrast, LLA takes the individual class-level regions as the processing unit and applies transformation respectively. The augmented regions are then combined linearly. This process can be represented as:
\begin{equation}
{\rm LLA}(x,m) =\textstyle\sum_{c=1}^{C} \alpha_c \mathcal{F}_{{\rm p}_c}(x^c) + \beta_c,
\label{eq:lla}
\end{equation}
where $\alpha_c \sim\mathcal{TN}(1,\sigma_1)$ and $\beta_c\sim\mathcal{TN}(0,\sigma_2)$ are {location-scale factors}. For all classes, ${\rm p}_c = 0.5$ is set to apply random inversion, except ${\rm p}_1 = 1$ is set to ensure the LLA augmented images are dissimilar to the GLA samples. 
It should be noted that we only perform non-linear transformation operations on foreground regions (the non-blank regions in the pictures, representing the human body areas). We visualize the GLA and LLA process in Fig.~\ref{fig:bezier} for better understanding. Empirically, $\sigma_1$ is set to be small (e.g. $0.1$ used in the paper), encouraging that the augmented data should stay not further from the source data; $\sigma_2$ is set to $0.5$ in this paper, constraining that the ``shift" is neither too small nor too big.

\subsection{Theoretical Evidence}
We provide the theoretical evidence to show that our framework can lead to an upper bound of the expected loss on “unseen" but related target domain under our setting.

\newtheorem{assumption}{Assumption}
\newtheorem{proposition}{Proposition}
\newtheorem{theorem}{Theorem}

\begin{assumption}
The unseen target data can be represented by a linear combination of $C$ (the class number) random variables, each following the location-scale distribution of the source,  i.e., {$x_U =  \sum_{c=1}^{C}\alpha^c x_S^c + \beta^c$}.

\end{assumption}

\noindent Note that Assumption 1 is very mild  in the field of medical imaging tasks based on the observation that medical image domain variability is mainly about the overall contrast, texture and clarity at class-level, while other characteristics are comparably more consistent.

\begin{assumption}
There exists a distribution gap between the generated samples from different augmentations, i.e., $\phi^{aug{'}}_{S}\neq \phi^{aug{''}}_{S}$. The $\mathcal{H}$-divergence measured between the two distributions is bounded by $\delta$, which is the largest $\mathcal{H}$-divergence measured between elements of $\sum_{i=1}^{N}\phi^{aug^{i}}_{S}$, i.e., $d_{\mathcal{H}}[{\phi^{aug{'}}_{S}},{\phi^{aug{''}}_{S}}]\leq 
\delta$.
\end{assumption}

\noindent This assumption is satisfied in our setting as the processing steps vary for two augmentation strategy (GLA\&LLA), and the variation is bounded by the {semantic consistency}.

\begin{proposition}
\textbf{\rm{[\citet{albuquerque2019generalizing}]}} Define the convex hull $\Lambda_S$ as the set of source mixture distributions $\Lambda_S = \left\{ {\bar{\phi} : \bar{\phi}(\cdot) = \sum_{i=1} ^{N_S}} \pi_{i} \phi_{S}^{i}(\cdot),\pi_{i}\in\delta_{N_S-1}\right\}$, where $\delta_{N_S-1}$ is the $N_S$-1-th dimensional simplex.
The unseen domain composed on a set of unseen distributions $\phi_{U}^{j}, j\in [N_U]$ is defined as $\phi_{U}$. The labeling rules corresponding to each domain are denoted as $f_{S_i}$ and $f_{U_i}$. We further introduce $\bar{\phi}_{U}$, the element within $\Lambda_S$ which is closest to $\phi_{U}$, i.e., $\bar{\phi}_{U}$ is given by $\mathop{\arg\min}_{\pi_{1},...,\pi_{N_{S}}}d_\mathcal{H}[\phi_{U},\sum_{i=1}^{N_S}\pi_{i}\phi_{S}^{i}]$. The risk $R_U[h], \forall h \in \mathcal{H},$ for any unseen domain $\phi_U$ such that $d_{\mathcal{H}}[\bar{\phi}_U,{\phi}_U]=\gamma$ is bounded as:
\end{proposition}
\begin{equation}
\small
\begin{split}
R_{U}[h] &\leq \textstyle\sum_{i=1}^{N_S}\pi_{i}R_{S}^{i}[h] + \gamma +{\delta} \\&+ min\left\{\mathbb{E}_{\bar{\phi}_{U}}[|f_{s_{\pi}}-f_U|],\mathbb{E}_{{\phi}_{U}}[|f_U-f_{s_{\pi}}|]\right\} ,
\end{split}
\label{proposition}
\end{equation}
\noindent where $\delta$ is the highest pairwise $\widehat{\mathcal{H}}$-divergence within $\Lambda_S$, \\
\noindent$\widehat{\mathcal{H}} = \left\{sign(|h(x)-h^{'}(x)|-t)|h, h^{'}(x)\in \mathcal{H},0\leq t\leq 1\right\}$.
$f_{S_{\pi}}(x) = \sum_{i=1} ^{N_S} \pi_{i}f_{S_i}(x)$ is the labeling function for any $x \in Supp(\bar{\phi}_U)$ resulting from combining all $f_{S_i}$ with weight $\pi_{i}$, $i\in[N_S], \sum_{i=1}^{N_S}\pi_{i}=1$, determined by $\bar{\phi}_U$.

\begin{theorem}
Based on Assumption 1 \& 2, given data from $N_S$ source domains, where the empirical risk of domain i is given as $L(y^i,y) = \epsilon_{i}\leq\sigma$, the single-source domain generalization risk can be upper bounded by $\sigma + \delta$. 
\end{theorem}

\noindent\textit{Proof.}  In our task -  single-source domain generalization, the \textit{covariate shift assumption}~\citep{david2010impossibility} holds,  the labeling function remains unchanged. When such assumption holds, $f_{S_{\pi}}=f_{{U}}$. The right-most term in Eqn.~\ref{proposition} equals to zero.
{Based on Assumption 1, the unseen target data can be represented by a linear combination of $C$ random variables sampled from location-scale distribution of the source. Accordingly, data augmentation is implemented with random sampling from the assumed distribution. This can lead to the conclusion that the unseen distribution is actually contained inside the convex hull of the sources, i.e.,}  $\phi_U \in \Lambda_S, \phi_U=\bar{\phi}_U,  d_{\mathcal{H}}[\bar{\phi}_U,{\phi}_U]= \gamma =0$.
The upper bound would be rewritten as:
\begin{equation}
\small
\begin{split}
R_{U}[h] \leq \textstyle\sum_{i=1}^{N_S}\pi_{i}R_{S}^{i}[h] + \delta
\leq \sigma + \delta.
\end{split}
\label{eq:upperbound}
\end{equation}
This completes the proof. \hfill $\Box$

Notice that the upper bound only depends on the source risk $\sigma$ and the source distribution divergence $\delta$. $\sigma$ is minimized during the training process. As illustrated in Assumption 2, $\delta$ is bounded by the semantic consistency, which means $\delta$ is low in actual experiments. We further validate that the saliency-balancing fusion would lead to a tighter bound compared with no fusion and random fusion through experiments in the following parts.

\begin{algorithm}[tb]
\caption{\small Saliency-balancing Location-scale Augmentation}
\label{alg:algorithm} 
\textbf{Require}: Training data $(x, m)$, network $f_\theta$, loss function $\mathcal{L}$, global location-scale augmentation ${\rm GLA}$, local location-scale augmentation ${\rm LLA}$, {common augmentation $F$}.

\begin{algorithmic}[1] 
\STATE $x^g  \leftarrow {\rm GLA}(x)$   \hfill  {$\triangleright$ See Eqn.\ref{eq:gla}}
\STATE $x^l \leftarrow {\rm LLA}(x, m)$ \hfill {$\triangleright$ See Eqn.\ref{eq:lla}}
\STATE $\tilde{x}^g, \tilde{x}^l, \tilde{m}  \leftarrow F(x^g, x^l,m)$ \hfill{ $\triangleright$ Common augmentation}
\STATE Calculate gradient \textit{Grad} = $\nabla_{\tilde{x}^g}\mathcal{L}(f_\theta(\tilde{x}^g),\tilde{m})$
\STATE $\textbf{s} \leftarrow \textit{normalize}(\textit{smooth}(\left|\textit{Grad}\right|$)) \hfill {$\triangleright$ Calculate saliency}
\STATE $\tilde{x}^{fused}  \leftarrow\textbf{s} \odot \tilde{x}^g+ (1-\textbf{s}) \odot \tilde{x}^l$
\STATE \textbf{return} $\tilde{x}^{fused}$
\end{algorithmic}
\end{algorithm}

\subsection{Saliency-balancing Fusion}
We further introduce a gradient distribution harmonizing mechanism named Saliency-balancing Fusion (SBF). 
The main idea is to preserve the sensitive areas (large-gradient areas) while progressively driving the span of the training data, enabling the augmentation progress to evolve under the guidance of gradient information.
Specifically, the saliency map is obtained by taking $l_2$ norm of gradient values across input channels and then downsampled to the grid size of $g \times g$ followed by interpolation to the original image size via quadratic B-spline kernels for smoothing.

The whole process can be described as follows: First, we input a GLA augmented image and calculate the saliency map. This step will identify the sensitive areas in source images with slight distribution shifts.
Second, we fuse LLA augmented image with GLA augmented image through saliency map, which preserves the large-gradient regions in the GLA augmented image and replace the remaining regions with LLA augmented parts, where drastic changes could be observed in the appearance. Algorithm 1 describes the whole process of our augmentation pipeline, while the above mentioned SBF is detailed in line 4-6.
As the fused images contain both parts of the augmented images, SBF encourages the model to behave linearly in-between GLA augmented and LLA augmented examples. We argue that this linear behaviour reduces the amount of undesirable oscillations when predicting  out-of-distribution examples compared with directly inputting LLA augmented samples. 

Finally, we utilize both GLA augmented images and the saliency-balancing fused images to train the model. The segmentation model $f_\theta$ is optimized with the total objective consisting of the cross entropy loss and Dice loss as follows:
\begin{equation}
\mathcal{L}(x,m) = \mathcal{L}_{ce}(x,m)+\mathcal{L}_{dice}(x,m).
\end{equation}

\section{Experiments and Results}

\subsection{Datasets and Preprocessing}
In our experiments, we evaluate our method on two datasets, cross-modality abdominal dataset~\citep{abdominalCT,abdominalMRI} and cross-sequence cardiac dataset~\citep{zhuang2020cardiac}. The detailed split of dataset and the preprocessing steps follow the instructions given by ~\citet{ouyang2020self}, which can be found in the given code.
Our proposed approach is used as additional stages followed by common augmentations including Affine, Elastic, Brightness, Contrast, Gamma and Additive Gaussian Noise. All the compared methods (including ``ERM'' and ``Supervised'') conduct the same common augmentation for fair comparison.

\subsection{Network Architecture and Training Configurations}
We utilize U-Net with an EfficientNet-b2 backbone as our segmentation network, same as CSDG~\citep{CSDG}. The network is trained from scratch. The grid size $g$ is empirically set to {3 and 18 for abdominal and cardiac datasets respectively}. Adam~\citep{adam} is used as the optimizer with an initial learning rate of $3\times 10^{-4}$ and weight decay of $3\times 10^{-5}$. The learning rate remains unchanged for the first 50 epochs and linearly decays to zero over the next 1,950 epochs. For all experiments, batch size is set to 32 and the methods are evaluated at the 2,000$^{th}$ epoch. We implemented our framework on a workstation equipped with one NVIDIA GeForce RTX 3090 GPU (24G memory).

\begin{table*}[h]
\small
\centering
\begin{tabular}{c|ccccc|cccc}\hline
\multirow{2}{*}{Method} & \multicolumn{5}{c|}{Abdominal CT-MRI} & \multicolumn{4}{c}{Cardiac bSSFP-LGE} \\
\cline{2-10}
 & Liver & R-Kidney & L-Kidney & \multicolumn{1}{c|}{Spleen} & Average & LVC & MYO & \multicolumn{1}{c|}{RVC} & Average \\ \hline
Supervised & 91.30 & 92.43 & 89.86 & \multicolumn{1}{c|}{89.83} & 90.85 & 92.04 & 83.11 & \multicolumn{1}{c|}{89.30} & 88.15 \\ 
ERM & 78.03 & 78.11 & 78.45 & \multicolumn{1}{c|}{74.65} & 77.31 & 86.06 & 66.98 & \multicolumn{1}{c|}{74.94} & 75.99 \\ \hline
Cutout & 79.80 & 82.32 & 82.14 & \multicolumn{1}{c|}{76.24} & 80.12 & 88.35 & 69.06 & \multicolumn{1}{c|}{79.19} & 78.87 \\
RSC & 76.40 & 75.79 & 76.60 & \multicolumn{1}{c|}{67.56} & 74.09 & 87.06 & 69.77 & \multicolumn{1}{c|}{75.69} & 77.51 \\
MixStyle & 77.63 & 78.41 & 78.03 & \multicolumn{1}{c|}{77.12} & 77.80 & 85.78 & 64.23 & \multicolumn{1}{c|}{75.61} & 75.21 \\
AdvBias & 78.54 & 81.70 & 80.69 & \multicolumn{1}{c|}{79.73} & 80.17 & 88.23 & 70.29 & \multicolumn{1}{c|}{80.32} & 79.62 \\
RandConv & 73.63 & 79.69 & 85.89 & \multicolumn{1}{c|}{83.43} & 80.66 & 89.88 & 75.60 & \multicolumn{1}{c|}{85.70} & 83.73 \\
CSDG & \underline{86.62} & \underline{87.48} & \underline{86.88} & \multicolumn{1}{c|}{\underline{84.27}} & \underline{86.31} & \underline{90.35} & \underline{77.82} & \multicolumn{1}{c|}{\underline{86.87}} & \underline{85.01} \\ \hline
SLAug (ours) & \textbf{90.08} & \textbf{89.23} & \textbf{87.54} & \multicolumn{1}{c|}{\textbf{87.67}} & \textbf{88.63} & \textbf{91.53} & \textbf{80.65} & \multicolumn{1}{c|}{\textbf{87.90}} & \textbf{86.69} \\ \hline\hline
\multirow{2}{*}{Method} & \multicolumn{5}{c|}{Abdominal MRI-CT} & \multicolumn{4}{c}{Cardiac LGE-bSSFP} \\
\cline{2-10}
 & Liver & R-Kidney & L-Kidney & \multicolumn{1}{c|}{Spleen} & Average & LVC & MYO & \multicolumn{1}{c|}{RVC} & Average \\ \hline
Supervised & 98.87 & 92.11 & 91.75 & \multicolumn{1}{c|}{88.55} & 89.74 & 91.16 & 82.93 & \multicolumn{1}{c|}{90.39} & 88.16 \\ 
ERM & 87.90 & 40.44 & 65.17 & \multicolumn{1}{c|}{55.90} & 62.35 & 90.16 & 78.59 & \multicolumn{1}{c|}{87.04} & 85.26 \\ \hline
Cutout & 86.99 & 63.66 & 73.74 & \multicolumn{1}{c|}{57.60} & 70.50 & 90.88 &79.14  & \multicolumn{1}{c|}{87.74} & 85.92 \\
RSC & \underline{88.10} & 46.60 & 75.94 & \multicolumn{1}{c|}{53.61} &66.07  & 90.21 & 78.63 & \multicolumn{1}{c|}{87.96} &85.60  \\
MixStyle & 86.66 & 48.26 & 65.20 & \multicolumn{1}{c|}{55.68} & 63.95 &  91.22&79.64  & \multicolumn{1}{c|}{88.16} & 86.34 \\
AdvBias & 87.63 & 52.48 & 68.28 & \multicolumn{1}{c|}{50.95} & 64.84 &91.20  &  79.50& \multicolumn{1}{c|}{88.10} &86.27  \\
RandConv & 84.14 & 76.81 & 77.99 & \multicolumn{1}{c|}{67.32} & 76.56 & \textbf{91.98} & \underline{80.92} & \multicolumn{1}{c|}{88.83} &\underline{ 87.24 }\\
CSDG & 85.62 & \underline{80.02} & \underline{80.42} & \multicolumn{1}{c|}{\underline{75.56}} & \underline{80.40} &91.37  & 80.43 & \multicolumn{1}{c|}{\underline{89.16}} & 86.99 \\ \hline
SLAug (ours) & \textbf{89.26} & \textbf{80.98} & \textbf{82.05} & \multicolumn{1}{c|}{\textbf{79.93}} & \textbf{83.05} & \underline{91.92} &\textbf{ 81.49} & \multicolumn{1}{c|}{\textbf{89.61}} & \textbf{87.67}\\  \hline
\end{tabular}
\caption{Performance comparison of different methods. Dice score (\%) is utilized as the evaluation metrics.}
\label{tab:compare}
\end{table*}

\begin{figure*}[h]
\centering
\includegraphics[width=1.8\columnwidth]{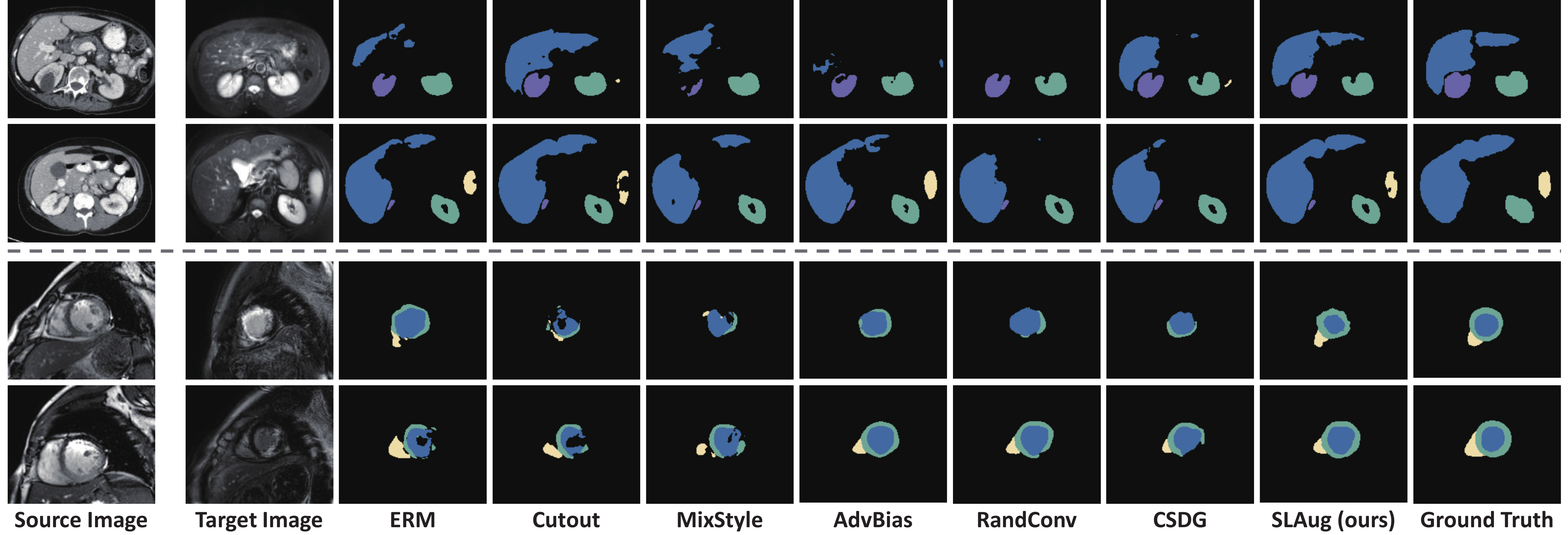}
\caption{Qualitative comparison for Abdominal CT-MRI (top two rows) and Cardiac bSSFP-LGE (bottom two rows).} 
\label{fig:compare}
\end{figure*}

\subsection{Results and Comparative Analysis}
To quantify the instance segmentation performance of each method, we engage the Dice score~\cite{dice} as the evaluation metric for measuring the overlap between the prediction and  ground truth. 

We compare our proposed method with the baseline empirical risk minimization (ERM) and the state-or-the-art methods (e.g. Cutout, RSC, MixStyle, AdvBias, RandConv, and CSGD) in the abdominal and cardiac segmentation. 
Cutout~\citep{cutout} improves the model robustness and overall performance by randomly masking out square regions.
RSC~\citep{RSC} forces the model to learn more general features by iteratively discarding the dominant features. 
MixStyle~\citep{mixstyle} generates novel images by mixing styles of training instances at the bottom layers of the network.
AdvBias~\citep{advbias} generates plausible and realistic signal corruptions, which models the intensity inhomogeneities in an adversarial way.
To randomize the image intensity and texture, RandConv~\citep{randconv} employs transformation via randomly initializing the weight of the first convolution layer.
CSDG~\citep{CSDG} further improves RandConv by extending the random module to a shallow network, and then mixes two augmented images through the pseudo correlations map to alleviate the spurious spatial correlations.

The quantitative performance of different methods is presented in Tab.~\ref{tab:compare}. Overall, our method significantly outperforms the previous methods by a large margin.
To be specific, we improve the performance of SDG by  47.77\% compared with the latest work CSDG in the four experiments on average, greatly narrowing the gap between single-source domain generalization and the upper bound. 
In abdominal cross-modality experiments, our method achieves the highest average Dice score of 88.63 and 83.05 in both directions. 
Compared with the most competitive method CSDG which engages a random texture augmentation, SLAug leads to an average improvement of 2.49 Dice score, demonstrating its superiority in cross-modality scenarios. 
In cardiac cross-sequence experiments, we observe the unequal difficulty between the generalization direction of ``bSSFP to LGE" and ``LGE to bSSFP". Our method performs steadily in either small or large domain gap situation, leading to an increase of 1.68 and 0.43 compared with the best performance in previous methods. 
This indicates that SLAug has better domain generalization capability in various medical image distributions with diverse domain gaps.
The visualization of different methods is presented in Fig.~\ref{fig:compare}. We also show the source and target domain images in the first two columns, illustrating the appearance shift between domains.  It is clearly observed that our method generates few misclassified predictions in the unseen target domain.

\begin{figure*}[h]
\centering
\includegraphics[width=1.9\columnwidth]{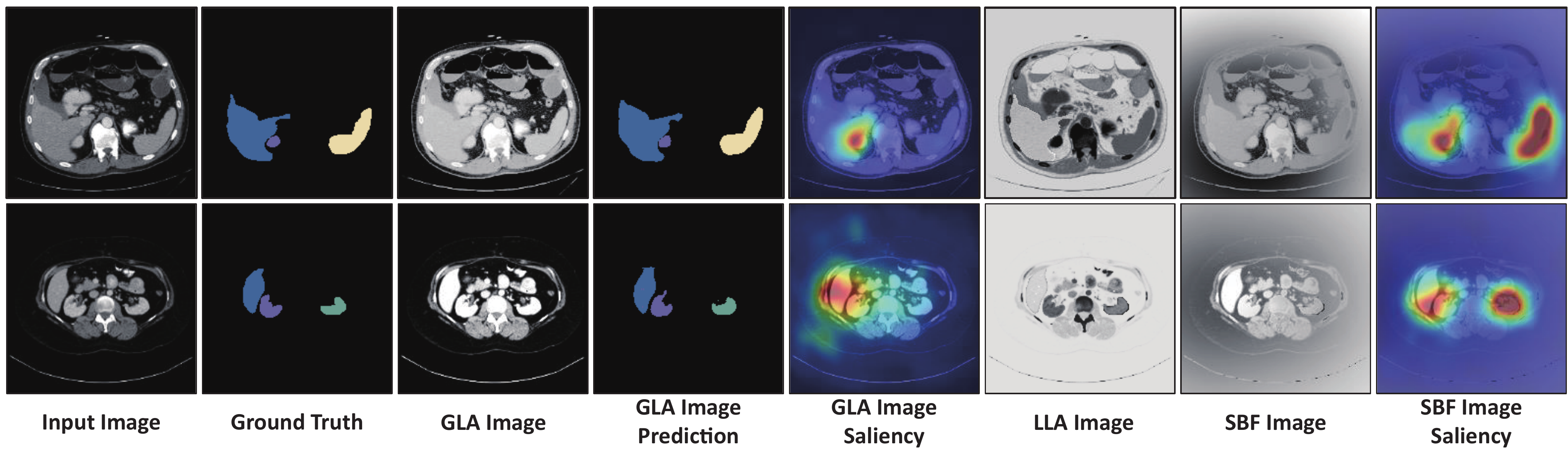}
\caption{Visualization of the saliency map from intermediate results of SLAug. }
\label{fig:ablation}
\end{figure*}

\subsection{Analytical Experiments}
\paragraph{Ablation Study.} We conduct ablation studies on the three main components in SLAug to better demonstrate our contribution, i.e., the Global Location-scale Augmentation~(GLA), the Local Location-scale Augmentation~(LLA), and Saliency-balancing Fusion~(SBF). Tab.~\ref{tab:abla} presents the performance of different variants of SLAug in two tasks - abdominal cross-modality and cardiac cross-sequence segmentation.
Comparing Variant 1, 2 with ERM, we can observe that solely utilizing GLA or LLA could steadily improve the overall performance.
However, GLA and LLA can only boost the performance in one of the two tasks, indicating either GLA or LLA is not applicable for both types of tasks.
Variant 2 indicates that  utilizing mask-based augmentation could lead to a gain in performance,  but once GLA is combined (e.g. Variant 3),  the obtained results could even outperform most SOTA methods. 
Then, we add SBF based on Variant 1,2 and 3. For Variant 4 and 5, we augment the images with the same augmentation twice, and use one of them to generate the saliency map to fuse the two images.
The performance improvement of Variant 4 over Variant 1 implies that the saliency information from GLA augmented images plays a vital role in our method, as it can indicate the data distribution near the decision boundary of the source domain, thus providing guidance for distribution enlargement.
Also, the marginal performance degradation in Variant 5 over Variant 2 also supports this claim. We assume that the saliency information derived from LLA augmented samples is inaccurate since LLA samples vary a lot from the original source samples, which leads to the decrease of the overall performance.
Finally, SLAug performs the best, showing that the three components promote mutually and are all indispensable for the superior domain generalization.

\begin{table}[t]
\centering
\small
\begin{tabular}{c|ccc|cc|c}\toprule
Methods & GLA & LLA & SBF & Abd. & Card. & Avg. \\\hline
ERM & - & - & - & 77.31 & 75.99& 76.65\\
Variant 1 & \checkmark & - & - & 80.28 & 74.21 & 77.25 \\
Variant 2 & - & \checkmark & - & 75.43 & 85.12 & 80.28 \\
Variant 3 & \checkmark & \checkmark & - & 85.48 & 85.55 & 85.52 \\
Variant 4 & \checkmark & - & \checkmark & 83.57 & 82.39 & 82.98 \\
Variant 5 & - & \checkmark & \checkmark & 75.42 & 85.00 & 80.21 \\
SLAug & \checkmark & \checkmark & \checkmark & \textbf{88.63} & \textbf{86.69} & \textbf{87.66}\\\bottomrule
\end{tabular}
\caption{Ablation study of SLAug on the direction of abdominal CT-MRI (Abd.) and cardiac bSSFP-LGE (Card.).}
\label{tab:abla}
\end{table}
\begin{figure}[t]
\centering
\includegraphics[width=1\columnwidth]{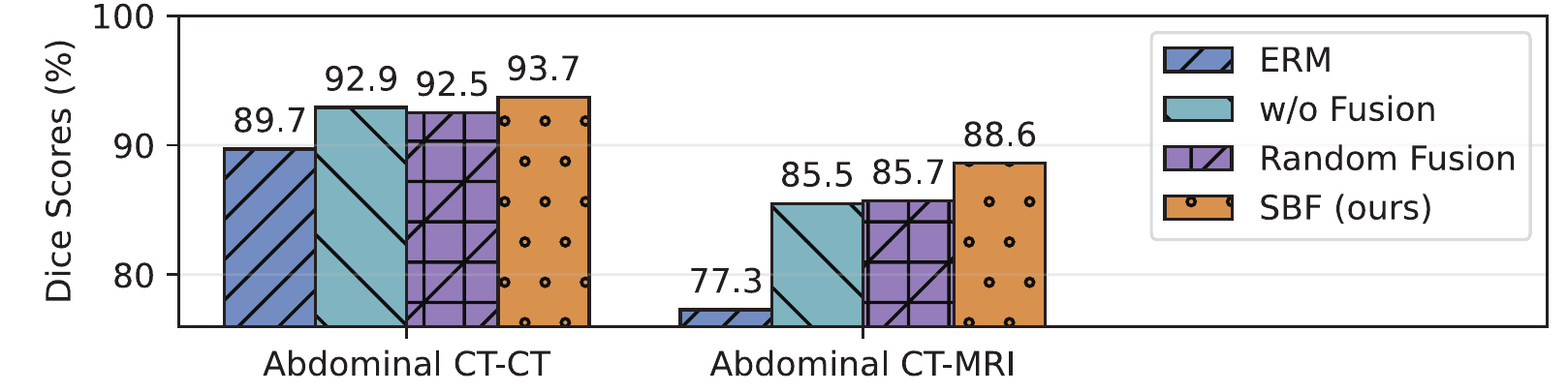}
\caption{Comparisons of ERM and different fusion methods on location-scale augmented images in Abdominal task.}
\label{fig:sal}
\end{figure}

\paragraph{Effectiveness of Saliency-balancing Fusion.}
To investigate the benefits of the guidance from saliency for image fusion, we replace the saliency map with a randomly initialized map. As shown in Fig.~\ref{fig:sal}, random fusion slightly improves the performance by 0.2 Dice score against without (w/o) fusion in cross-domain (CT-MRI) scenarios. This indicates the mixture of two augmented images could improve the generalization performance, which, may benefit from the enrichment of the augmented samples.
However, this result is much lower than SBF by 2.9 in Dice score, as we consider SBF is much more informative and explainable compared with random fusion.
Moreover, we compare the source test set results as a metric for the preserving capability of source-domain knowledge, as shown in Fig.~\ref{fig:sal} by the CT-CT bars. A reduction of 0.4 Dice score can be observed by using random fusion against directly inputting the two augmented images into training (w/o fusion). In contrast, our SBF promotes the source performance by 0.8 in comparison with no fusion, and 4.0 compared with empirical risk minimization (ERM). This result corresponds to the first term of the upper bound in Eqn.~\ref{eq:upperbound}, 
showing that SBF could achieve a tighter generalization bound  by lowering the source risk.

Besides demonstrating the effectiveness of SBF by numerical values, we provide the visualization of the saliency maps and the predictions of the GLA augmented image, serving as the indicator for its interpretability and informativeness in Fig.~\ref{fig:ablation}. From these pictures, we can infer that 
1) saliency map can provide accurate feedback on the sensitive regions in GLA step. If we focus on the large-gradient areas in saliency map, we would find that these areas are misclassified in GLA prediction.
2) LLA enforces larger augmentation intensity on the given samples. Saliency maps in the last column show more salient areas in comparison to the GLA images saliency maps.
3) SBF could effectively combine the {sensitive areas} in GLA images while introduce larger augmentation magnitude in the remaining areas. 
The fused images can locate the vital areas, which is beneficial for gradient harmonizing, so as to boost the generalization capability of the training model.

\renewcommand{\thefootnote}{}

\section{Conclusion}
In this paper, we propose a novel augmentation strategy to better solve the single-source domain generalization problems in medical image segmentation.  By jointly integrate the location-scale augmentation and saliency-balancing fusion, this plug-and-play module shows superior performance on two challenging tasks.
Moreover, we provide the theoretical evidence and visualization examples for better explanation, aiming to facilitate more principled techniques for robust learning in medical fields. \footnote{This research was funded by National Natural Science Foundation of China under no.61876154, no.61876155 and no.62206225;  Jiangsu Science and Technology Programme (Natural Science Foundation of Jiangsu Province) under no. BE2020006-4; Natural Science Foundation of the Jiangsu Higher Education Institutions of China under no. 22KJB520039; Xi’an Jiaotong-Liverpool University's Key Program Special Fund under no. KSF-E-37 and no. KSF-T-06; and Research Development Fund in XJTLU under no.RDF-19-01-21.}

\bibliography{aaai23}

\begin{thebibliography}{33}
\providecommand{\natexlab}[1]{#1}

\bibitem[{Albuquerque et~al.(2019)Albuquerque, Monteiro, Darvishi, Falk, and
  Mitliagkas}]{albuquerque2019generalizing}
Albuquerque, I.; Monteiro, J.; Darvishi, M.; Falk, T.~H.; and Mitliagkas, I.
  2019.
\newblock Generalizing to unseen domains via distribution matching.
\newblock arXiv:1911.00804.

\bibitem[{Buzug(2011)}]{buzug2011computed}
Buzug, T.~M. 2011.
\newblock Computed tomography.
\newblock In \emph{Springer handbook of medical technology}, 311--342.
  Springer.

\bibitem[{Chen et~al.(2019)Chen, Dou, Chen, Qin, and Heng}]{SIFA}
Chen, C.; Dou, Q.; Chen, H.; Qin, J.; and Heng, P.-A. 2019.
\newblock Synergistic image and feature adaptation: Towards cross-modality
  domain adaptation for medical image segmentation.
\newblock In \emph{AAAI Conference on Artificial Intelligence}, volume~33,
  865--872.

\bibitem[{Chen et~al.(2020)Chen, Qin, Qiu, Ouyang, Wang, Chen, Tarroni, Bai,
  and Rueckert}]{advbias}
Chen, C.; Qin, C.; Qiu, H.; Ouyang, C.; Wang, S.; Chen, L.; Tarroni, G.; Bai,
  W.; and Rueckert, D. 2020.
\newblock Realistic adversarial data augmentation for MR image segmentation.
\newblock In \emph{International Conference on Medical Image Computing and
  Computer-Assisted Intervention}, 667--677. Springer.

\bibitem[{Chen et~al.(2021)Chen, Ouyang, Zhu, and Agam}]{chen2021mask}
Chen, Y.; Ouyang, X.; Zhu, K.; and Agam, G. 2021.
\newblock Mask-based data augmentation for semi-supervised semantic
  segmentation.
\newblock arXiv:2101.10156.

\bibitem[{David et~al.(2010)David, Lu, Luu, and
  P{\'a}l}]{david2010impossibility}
David, S.~B.; Lu, T.; Luu, T.; and P{\'a}l, D. 2010.
\newblock Impossibility theorems for domain adaptation.
\newblock In \emph{Proceedings of the Thirteenth International Conference on
  Artificial Intelligence and Statistics}, volume~9, 129--136.

\bibitem[{DeVries and Taylor(2017)}]{cutout}
DeVries, T.; and Taylor, G.~W. 2017.
\newblock Improved regularization of convolutional neural networks with cutout.
\newblock arXiv:1708.04552.

\bibitem[{Forbes(2012)}]{forbes2012human}
Forbes, G.~B. 2012.
\newblock \emph{Human body composition: growth, aging, nutrition, and
  activity}.
\newblock Springer Science \& Business Media.

\bibitem[{Hou and Zhang(2007)}]{hou2007saliency}
Hou, X.; and Zhang, L. 2007.
\newblock Saliency detection: A spectral residual approach.
\newblock In \emph{2007 IEEE Conference on computer vision and pattern
  recognition}, 1--8. Ieee.

\bibitem[{Huang et~al.(2020)Huang, Wang, Xing, and Huang}]{RSC}
Huang, Z.; Wang, H.; Xing, E.~P.; and Huang, D. 2020.
\newblock Self-challenging improves cross-domain generalization.
\newblock In \emph{European Conference on Computer Vision}, 124--140. Springer.

\bibitem[{Kavur et~al.(2021)Kavur, Gezer, Bar{\i}{\c{s}}, Aslan, Conze, Groza,
  Pham, Chatterjee, Ernst, {\"O}zkan et~al.}]{abdominalMRI}
Kavur, A.~E.; Gezer, N.~S.; Bar{\i}{\c{s}}, M.; Aslan, S.; Conze, P.-H.; Groza,
  V.; Pham, D.~D.; Chatterjee, S.; Ernst, P.; {\"O}zkan, S.; et~al. 2021.
\newblock CHAOS challenge-combined (CT-MR) healthy abdominal organ
  segmentation.
\newblock \emph{Medical Image Analysis}, 69: 101950.

\bibitem[{Kim, Choo, and Song(2020)}]{kim2020puzzle}
Kim, J.-H.; Choo, W.; and Song, H.~O. 2020.
\newblock Puzzle mix: Exploiting saliency and local statistics for optimal
  mixup.
\newblock In \emph{International Conference on Machine Learning}, 5275--5285.
  PMLR.

\bibitem[{Kingma and Ba(2014)}]{adam}
Kingma, D.~P.; and Ba, J. 2014.
\newblock Adam: A Method for Stochastic Optimization.
\newblock arXiv:1412.6980.

\bibitem[{Landman et~al.(2015)Landman, Xu, Igelsias, Styner, Langerak, and
  Klein}]{abdominalCT}
Landman, B.; Xu, Z.; Igelsias, J.; Styner, M.; Langerak, T.; and Klein, A.
  2015.
\newblock Miccai multi-atlas labeling beyond the cranial vault-workshop and
  challenge.
\newblock In \emph{Proc. MICCAI Multi-Atlas Labeling Beyond Cranial
  Vault—Workshop Challenge}, volume~5, 12.

\bibitem[{Milletari, Navab, and Ahmadi(2016)}]{dice}
Milletari, F.; Navab, N.; and Ahmadi, S.-A. 2016.
\newblock V-net: Fully convolutional neural networks for volumetric medical
  image segmentation.
\newblock In \emph{International Conference on 3D Vision}, 565--571. IEEE.

\bibitem[{Mortenson(1999)}]{mortenson1999mathematics}
Mortenson, M.~E. 1999.
\newblock \emph{Mathematics for computer graphics applications}.
\newblock Industrial Press Inc.

\bibitem[{Olsson et~al.(2021)Olsson, Tranheden, Pinto, and Svensson}]{classmix}
Olsson, V.; Tranheden, W.; Pinto, J.; and Svensson, L. 2021.
\newblock Classmix: Segmentation-based data augmentation for semi-supervised
  learning.
\newblock In \emph{Proceedings of the IEEE/CVF Winter Conference on
  Applications of Computer Vision}, 1369--1378.

\bibitem[{Ouyang et~al.(2020)Ouyang, Biffi, Chen, Kart, Qiu, and
  Rueckert}]{ouyang2020self}
Ouyang, C.; Biffi, C.; Chen, C.; Kart, T.; Qiu, H.; and Rueckert, D. 2020.
\newblock Self-supervision with superpixels: Training few-shot medical image
  segmentation without annotation.
\newblock In \emph{European Conference on Computer Vision}, 762--780. Springer.

\bibitem[{Ouyang et~al.(2021)Ouyang, Chen, Li, Li, Qin, Bai, and
  Rueckert}]{CSDG}
Ouyang, C.; Chen, C.; Li, S.; Li, Z.; Qin, C.; Bai, W.; and Rueckert, D. 2021.
\newblock Causality-inspired Single-source Domain Generalization for Medical
  Image Segmentation.
\newblock arXiv:2111.12525.

\bibitem[{Simonyan, Vedaldi, and Zisserman(2014)}]{simonyan2013deep}
Simonyan, K.; Vedaldi, A.; and Zisserman, A. 2014.
\newblock Deep Inside Convolutional Networks: Visualising Image Classification
  Models and Saliency Maps.
\newblock In \emph{2nd International Conference on Learning Representations,
  {ICLR} 2014, Banff, AB, Canada, April 14-16, 2014, Workshop Track
  Proceedings}.

\bibitem[{Valanarasu et~al.(2021)Valanarasu, Oza, Hacihaliloglu, and
  Patel}]{med1}
Valanarasu, J. M.~J.; Oza, P.; Hacihaliloglu, I.; and Patel, V.~M. 2021.
\newblock Medical Transformer: Gated Axial-Attention for Medical Image
  Segmentation.
\newblock In de~Bruijne, M.; Cattin, P.~C.; Cotin, S.; Padoy, N.; Speidel, S.;
  Zheng, Y.; and Essert, C., eds., \emph{Medical Image Computing and Computer
  Assisted Intervention}, volume 12901, 36--46.

\bibitem[{Volk et~al.(2019)Volk, M{\"u}ller, Von~Bernuth, Hospach, and
  Bringmann}]{volk2019towards}
Volk, G.; M{\"u}ller, S.; Von~Bernuth, A.; Hospach, D.; and Bringmann, O. 2019.
\newblock Towards robust CNN-based object detection through augmentation with
  synthetic rain variations.
\newblock In \emph{2019 IEEE Intelligent Transportation Systems Conference},
  285--292.

\bibitem[{Wang and Dudek(2014)}]{wang2014fast}
Wang, B.; and Dudek, P. 2014.
\newblock A fast self-tuning background subtraction algorithm.
\newblock In \emph{Proceedings of the IEEE Conference on Computer Vision and
  Pattern Recognition Workshops}, 395--398.

\bibitem[{Wang et~al.(2020)Wang, Yu, Li, Yang, Fu, and Heng}]{wang2020dofe}
Wang, S.; Yu, L.; Li, K.; Yang, X.; Fu, C.-W.; and Heng, P.-A. 2020.
\newblock Dofe: Domain-oriented feature embedding for generalizable fundus
  image segmentation on unseen datasets.
\newblock \emph{IEEE Transactions on Medical Imaging}, 39(12): 4237--4248.

\bibitem[{Wei et~al.(2017)Wei, Feng, Liang, Cheng, Zhao, and
  Yan}]{wei2017object}
Wei, Y.; Feng, J.; Liang, X.; Cheng, M.-M.; Zhao, Y.; and Yan, S. 2017.
\newblock Object region mining with adversarial erasing: A simple
  classification to semantic segmentation approach.
\newblock In \emph{Proceedings of the IEEE conference on computer vision and
  pattern recognition}, 1568--1576.

\bibitem[{Xu et~al.(2021)Xu, Liu, Yang, Raffel, and Niethammer}]{randconv}
Xu, Z.; Liu, D.; Yang, J.; Raffel, C.; and Niethammer, M. 2021.
\newblock Robust and Generalizable Visual Representation Learning via Random
  Convolutions.
\newblock In \emph{International Conference on Learning Representations}.

\bibitem[{Yao et~al.(2022)Yao, Su, Huang, Yang, Sun, Hussain, and
  Coenen}]{yao2022novel}
Yao, K.; Su, Z.; Huang, K.; Yang, X.; Sun, J.; Hussain, A.; and Coenen, F.
  2022.
\newblock A novel 3D unsupervised domain adaptation framework for
  cross-modality medical image segmentation.
\newblock \emph{IEEE Journal of Biomedical and Health Informatics}.

\bibitem[{Zhang, Zhang, and Xu(2021)}]{objectaug}
Zhang, J.; Zhang, Y.; and Xu, X. 2021.
\newblock Objectaug: object-level data augmentation for semantic image
  segmentation.
\newblock In \emph{2021 International Joint Conference on Neural Networks
  (IJCNN)}, 1--8. IEEE.

\bibitem[{Zhao et~al.(2015)Zhao, Ouyang, Li, and Wang}]{zhao2015saliency}
Zhao, R.; Ouyang, W.; Li, H.; and Wang, X. 2015.
\newblock Saliency detection by multi-context deep learning.
\newblock In \emph{Proceedings of the IEEE conference on computer vision and
  pattern recognition}, 1265--1274.

\bibitem[{Zhou et~al.(2021)Zhou, Yang, Qiao, and Xiang}]{mixstyle}
Zhou, K.; Yang, Y.; Qiao, Y.; and Xiang, T. 2021.
\newblock Domain Generalization with MixStyle.
\newblock In \emph{International Conference on Learning Representations}.

\bibitem[{Zhou et~al.(2022)Zhou, Qi, Yang, Ni, and Shi}]{dualnormal}
Zhou, Z.; Qi, L.; Yang, X.; Ni, D.; and Shi, Y. 2022.
\newblock Generalizable Cross-modality Medical Image Segmentation via Style
  Augmentation and Dual Normalization.
\newblock In \emph{Proceedings of the IEEE/CVF Conference on Computer Vision
  and Pattern Recognition}, 20856--20865.

\bibitem[{Zhou et~al.(2019)Zhou, Sodha, Rahman~Siddiquee, Feng, Tajbakhsh,
  Gotway, and Liang}]{zhou2019models}
Zhou, Z.; Sodha, V.; Rahman~Siddiquee, M.~M.; Feng, R.; Tajbakhsh, N.; Gotway,
  M.~B.; and Liang, J. 2019.
\newblock Models genesis: Generic autodidactic models for 3d medical image
  analysis.
\newblock In \emph{International conference on medical image computing and
  computer-assisted intervention}, 384--393. Springer.

\bibitem[{Zhuang et~al.(2020)Zhuang, Xu, Luo, Chen, Ouyang, Rueckert, Campello,
  Lekadir, Vesal, RaviKumar et~al.}]{zhuang2020cardiac}
Zhuang, X.; Xu, J.; Luo, X.; Chen, C.; Ouyang, C.; Rueckert, D.; Campello,
  V.~M.; Lekadir, K.; Vesal, S.; RaviKumar, N.; et~al. 2020.
\newblock Cardiac segmentation on late gadolinium enhancement MRI: a benchmark
  study from multi-sequence cardiac MR segmentation challenge.
\newblock arXiv:2006.12434.

\end{thebibliography}

\newpage

\newpage

\renewcommand{\thefigure}{S\arabic{figure}}

\renewcommand{\thetable}{S\arabic{table}}
\appendix
\section{Appendix}

\subsection{Additive Experimental Details}
In our experiments, we evaluate our method on two datasets. The first cross-modality abdominal dataset includes 30 volumes of CT data~\citep{abdominalCT}, and 20 volumes of T2-SPIR MRI data~\citep{abdominalMRI}. 
We conduct a 70\%-10\%-20\% data split for training, validation and testing set on source domain. All the target domain images are used to evaluate the domain generalization performance, same in \citep{liu2020shape}.
The second dataset is cross-sequence cardiac segmentation dataset~\citep{zhuang2020cardiac}, which consists of 45 balanced steady-state free precession (bSSFP) and 45 late gadolinium enhanced (LGE) MRI volumes. Data split in direction ``bSFFP to LGE" is the same as in the challenge~\citep{zhuang2020cardiac}, and the training and test sets are interchanged in the opposite direction.

The preprocessing steps follow the instructions given by ~\citet{ouyang2020self}.
All the 3D volumes are spatial normalized and reformatted into 2D slices, and then cropped/resized to $192 \times 192$ on XY-plane. For the abdominal CT dataset, a window of [-275, 125]~\citep{ouyang2020self} is applied on Hounsfield values. For the MRI datasets, the top 0.5\% intensity histogram are clipped. Our proposed approach is used as additional stages followed by common augmentations including Affine, Elastic, Brightness, Contrast, Gamma and Additive Gaussian Noise. All the compared methods (including ``ERM'' and ``Supervised'') conduct the same common augmentation for fair comparison.
The details of common augmentations are included in Tab.~\ref{tab:aug}, which are applied to all methods by default. The parameters setting is the same with \citet{CSDG}. 

During training, all the augmentations applied to GLA and LLA image for one slice are in the same random states. The same geometric transformations such as Affine and Elastic transformations ensure the two augmented images can be fused without spatial difference. The same intensity augmentations including Brightness, Contrast and Gamma augmentations can maintain the appearance difference caused by location-scale augmentations.

\begin{table}[h]
\centering
\begin{tabular}{|l|l|}\hline
Augmentations & Parameters  \\\hline
Affine Transformation   & \begin{tabular}[c]{@{}l@{}}Rotate: 20\\Shift: 15\\Shear: 20\\Scale: [0.5, 1.5]\end{tabular}  \\\hline
Elastic Transformation & \begin{tabular}[c]{@{}l@{}}Alpha: 20\\Sigma: 5\end{tabular}  \\\hline
Brightness & Brightness: [-10,10]  \\\hline
Contrast  & Contrast: [0.6,1.5]  \\\hline
Gamma   & Gamma: [0.2, 1.8]   \\\hline
Additive Gaussian Noise & Noise std: 0.15\\\hline                                           
\end{tabular}
\caption{Parameters of common augmentations used in all experiments by default.}
\label{tab:aug}
\end{table}

\subsection{Hyper-parameter Analysis}
\begin{figure}[t]
\centering
\includegraphics[width=\columnwidth]{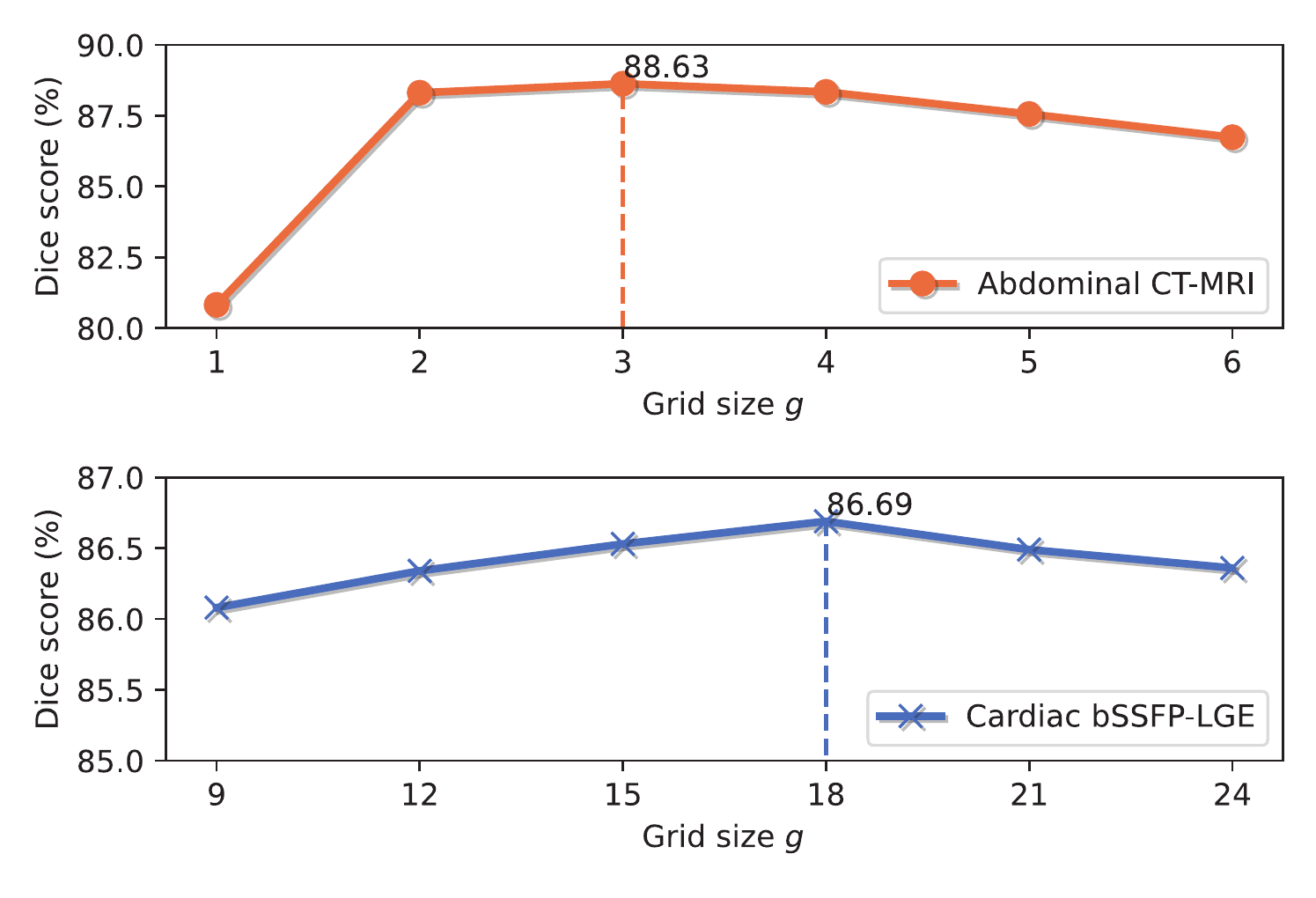}
\caption{Influence of grid size $g$ on two experiments.}
\label{fig:gride}
\end{figure}

We give an experimental analysis of the grid size $g$ as shown in Fig.~\ref{fig:gride},  which is utilized to smooth the gradient map. As can be observed in the figure, the performance remains steady with the change of grid size. Note that grid size $=1$ means the saliency map equals to 1 everywhere, so only GLA images contribute to the final fused image.

As the grid size affects the smoothness of the fine-grained fusion map, we conjecture that this parameter is related to the size and spatial distribution of each class. Different from segmenting separate organs in abdominal segmentation, the substructures in cardiac segmentation are more closer to each other. Therefore, a less smooth fusion map is expected in cardiac segmentation to ensure a higher resolution fusion map to distinguish each components inside the heart. Generally, the grid size is suggested to set large if the classes are close to each other.


\subsection{More Visualization}
\begin{figure*}[!h]
\centering
\includegraphics[width=1.6\columnwidth]{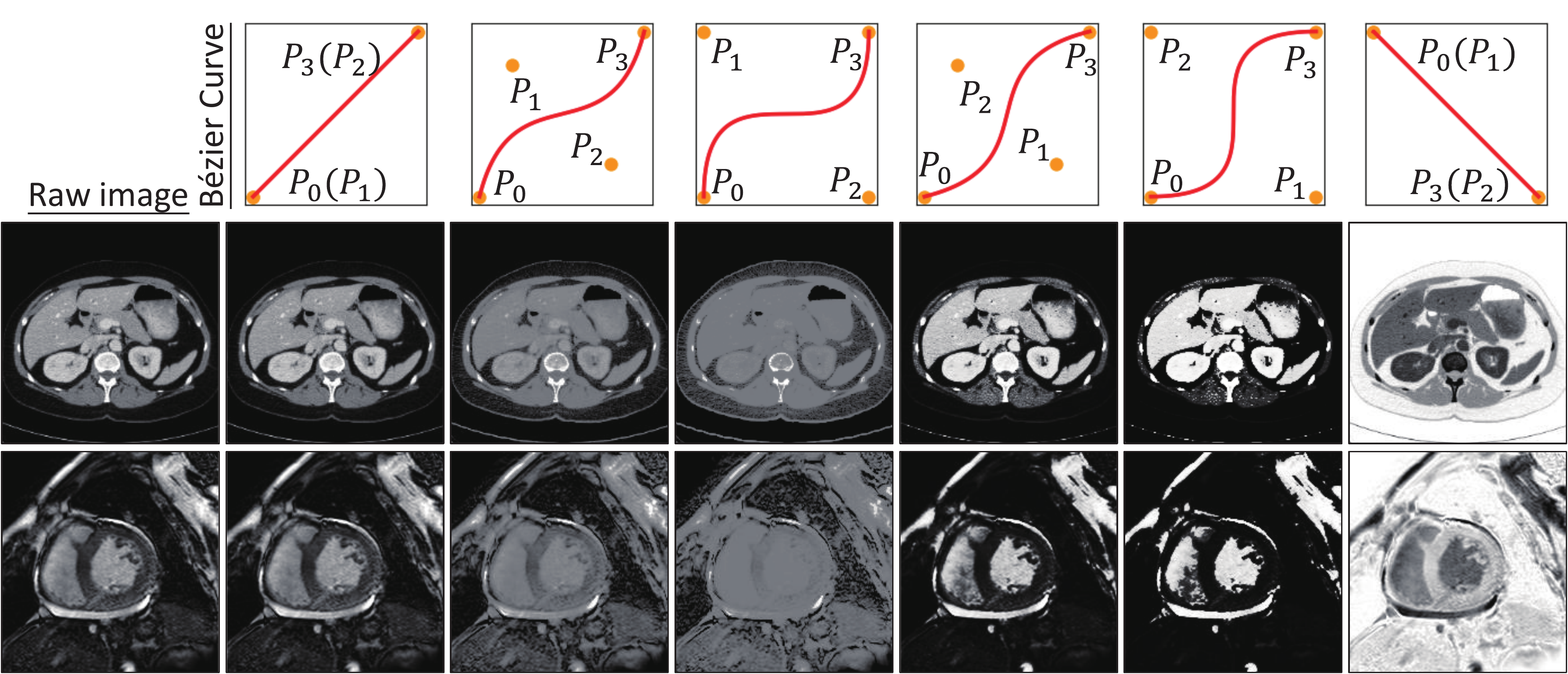}
\caption{Demonstration of bezier curve and augmented images.}
\label{fig:supbezier}
\end{figure*}

To better illustrate the Bezier Curve transformation, we demonstrate some cases in Fig.~\ref{fig:supbezier}. The 1st column is the raw images which have been normalized in $[0, 1]$. The start point $P_0=(0,0)$ and end point $P_3=(1,1)$ is set for simplicity, except $P_0=(0,1)$, $P_3=(1,0)$ is set at 7th columns for intensity inversion. 

\begin{figure*}[!h]
\centering
\includegraphics[width=1.8\columnwidth]{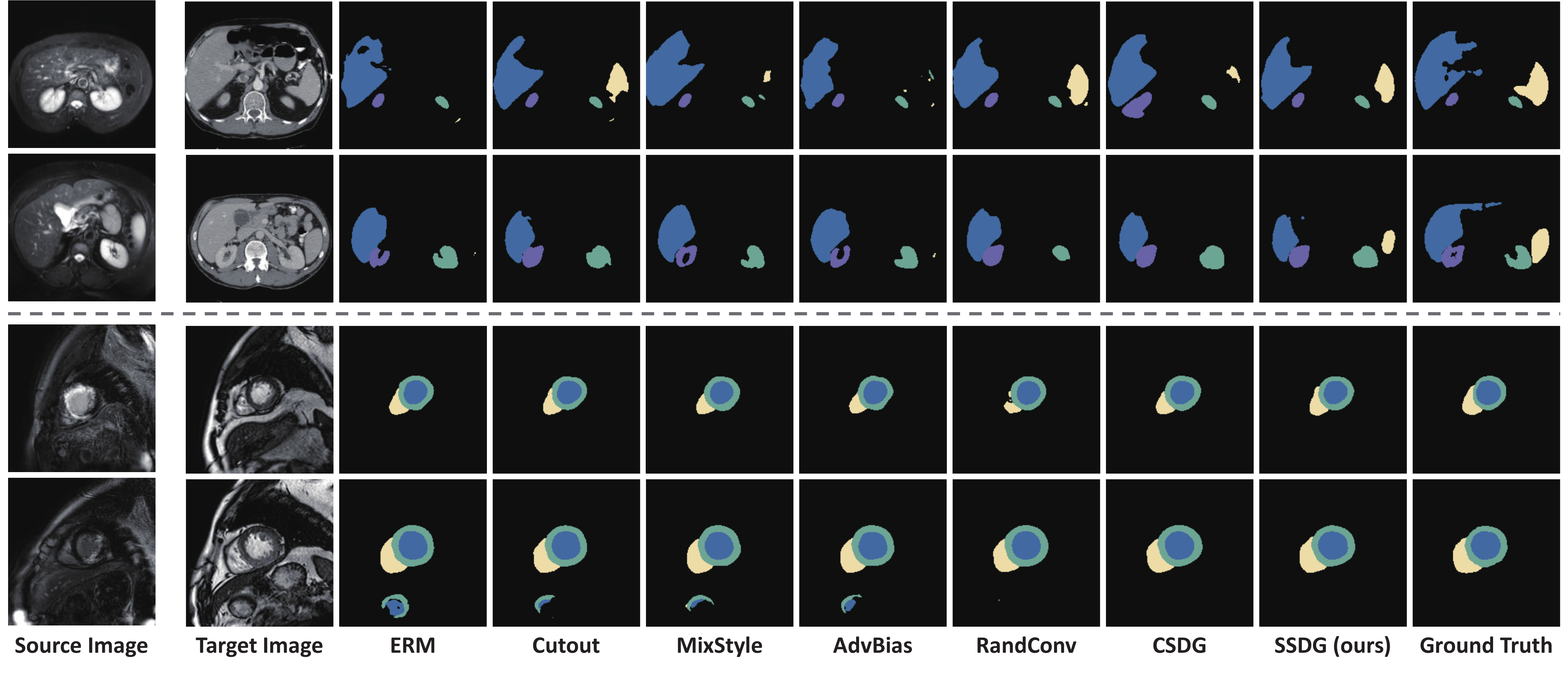}
\caption{Additive quantitative comparison for Abdominal MRI-CT (top two rows) and Cardiac LGE-bSSFP (bottom two rows).From left to right are the source domain images (1st column), the target domain images (2rd column), results of other comparative methods (3nd-8th columns), results of our proposed SLAug (9th column) and ground truth (last column).}
\label{fig:supcompare}
\end{figure*}

Meanwhile, we give more visualization of comparative results in Fig.~\ref{fig:supcompare}, consisting of Abdominal MRI-CT and Cardiac LGE-bSSFP segmentation.

\end{document}